\pdfoutput=1

\documentclass[11pt]{article}

\usepackage[]{naacl2021}

\usepackage{times}
\usepackage{latexsym}

\usepackage[T1]{fontenc}

\usepackage[utf8]{inputenc}

\usepackage{microtype}

\usepackage{mathtools}
\usepackage{amsfonts}
\usepackage{multirow}
\usepackage{xspace}
\usepackage{booktabs}

\newcommand{\simlength}{\textsc{SimiLe}\xspace}
\title{On Learning Text Style Transfer with Direct Rewards}

\author{Yixin Liu, Graham Neubig, John Wieting \\
  Carnegie Mellon University \\
  \texttt{\{yixinl2,gneubig,jwieting\}@cs.cmu.edu}
  }

\begin{document}
\maketitle
\begin{abstract}
In most cases, the lack of parallel corpora makes it impossible to directly train supervised models for the text style transfer task. In this paper, we explore training algorithms that instead optimize reward functions that explicitly consider different aspects of the style-transferred outputs. In particular, we leverage semantic similarity metrics originally used for fine-tuning neural machine translation models to explicitly assess the preservation of content between system outputs and input texts. We also investigate the potential weaknesses of the existing automatic metrics and propose efficient strategies of using these metrics for training. The experimental results show that our model provides significant gains in both automatic and human evaluation over strong baselines, indicating the effectiveness of our proposed methods and training strategies.\footnote{Code and data are available at: \url{https://github.com/yixinL7/Direct-Style-Transfer}}
\end{abstract}

\newcommand{\model}{\textsc{DirR}\xspace}
\newcommand{\modelbleu}{\textsc{DirR-Bleu}\xspace}
\newcommand{\modelcycle}{\textsc{DirR-Cycle}\xspace}
\newcommand{\modelsim}{\textsc{DirR} w/o \textsc{Flu}\xspace}
\section{Introduction}
Text style transfer aims to convert an input text into another generated text with a different style but the same basic semantics as the input.
One major challenge in this setting is that many style transfer tasks lack parallel corpora, 
since the absence of human references makes it impossible to train the text style transfer models using maximum likelihood estimation (MLE), which aims to maximize the predicted likelihood of the references. 
As a result, some of the earliest work~\citep{shen2017style,hu2017toward,fu2018style} on unsupervised text style transfer proposed training algorithms that are still based on MLE by formulating the style transfer models as auto-encoders optimized with reconstruction loss. 
Specifically, during training the model is tasked to generate a \emph{style-agnostic encoding} and reconstruct the input text based on this encoding with style-specific embeddings or decoders. 
During inference, the model aims to transfer the source text style using the target style information. 
While these methods have seen empirical success, they face the inherent difficulty of coming up with a style-agnostic but content-preserving encoding -- this is a non-trivial task and failure at this first step will diminish style transfer accuracy and content preservation of the final output.

Another line of work~\citep{xu2018unpaired, pang2019unsup, ijcai2019-0711} proposes training algorithms based on rewards related to the automatic evaluation metrics, which can assess the model performance more directly during training. 
This approach is conceptually similar to training algorithms that optimize models using rewards related to the corresponding evaluation metrics for other NLP tasks,  such as machine translation~\citep{shen-etal-2016-minimum, wieting-etal-2019-beyond} or text summarization~\citep{DBLP:conf/iclr/PaulusXS18, li-etal-2019-deep}.
As for unsupervised style transfer, the widely used automatic metrics mainly attend to three desiderata: (1) style transfer accuracy -- the generated sentence must be in the target style, commonly measured by the accuracy of a style classifier applied to the transferred text, (2) fluency -- the generated text must be grammatically correct and natural, commonly measured by the perplexity of a language model and (3) content preservation -- the semantics need to be preserved between the source and target, commonly measured by the BLEU score between the system outputs and source texts. 
Since these automatic metrics only require the system outputs and source texts, they can be used as rewards for training. 
Moreover, the two lines of approaches can be used together, and previous work~\citep{yang2018unsupervised,john-etal-2019-disentangled,madaan-etal-2020-politeness} proposed methods which use the auto-encoders as the backbone augmented with task-specific rewards. In particular, the style transfer accuracy reward is used by most of the recent work.

However, reward-based training algorithms still have their limitations, and in this paper we aim to identify and address the bottlenecks of these methods. 
Specifically, we focus on two problems: (1) the difficulty of designing an efficient reward for content preservation, (2) the lack of robustness of the existing automatic evaluation metrics.  

Content preservation is more difficult to measure compared to style transfer accuracy and fluency because it needs to consider the overlap in the semantics between the source text and system outputs.
While using BLEU score between the source text and system output would be a direct solution~\citep{xu2018unpaired}, this approach has an inherent limitation in that $n$-gram based metrics such as BLEU are sensitive to lexical differences and will penalize modifications that are necessary for transferring text style. 
In fact, previous work has proposed various different proxy rewards for content preservation. One of the most popular methods is the cycle-consistency loss~\citep{ijcai2019-0711, dai2019style, pang2019unsup}, which introduces a round-trip generation process, where the model generates an output in the target style, and the ability of a reconstruction model to re-generate the original text is used as a proxy for content preservation. 
While this method is more tolerant to lexical differences, the correlation between the reconstruction loss and content preservation can be weak.

Therefore, we aim to design a reward for content preservation which can directly assess the semantic similarity \emph{between the system outputs and input texts}. Specifically, we note that models of semantic similarity are widely studied \cite{DBLP:journals/corr/WietingBGL15a, sharma2017nlgeval, pagliardini-etal-2018-unsupervised, Zhang*2020BERTScore:},
and we can leverage these methods to directly calculate the similarity between the system outputs and input texts.
This renders our method applicable for even unsupervised settings where no human references are available.

Another key challenge for reward-based training algorithms is that the existing automatic evaluation metrics are not well-correlated with human evaluation~\citep{li-etal-2018-delete}. 
It poses general risks to the work in this field with respect to model training and evaluation since these metrics are widely used.  
An important observation we made from our experiments is that style transfer models can exploit the weaknesses of the automatic metrics. They do this by making minimal changes to the input texts which are enough to trick the classifier used for style transfer accuracy while achieving high content preservation and fluency scores due to the high lexical similarity with the input texts. 
Upon identifying this risk, we re-visit and propose several strategies that serve as auxiliary regularization on the style transfer models, effectively mitigating the problem discussed above.

We empirically show that our proposed reward functions can provide significant gains in both automatic and human evaluation over strong baselines from the literature. 
In addition, the problems we identify with existing automatic evaluation metrics suggest that the automatic metrics need to be used with caution either for model training or evaluation in order to make it truthfully reflect human evaluation.

\section{Methods}
\subsection{Overview}
Data for unsupervised text style transfer can be defined as $$D = \{(x^{(1)}, s^{(1)}), ..., (x^{(i)}, s^{(i)}), ..., (x^{(n)}, s^{(n)})\}, $$
where $x^{(i)}$ denotes the text and $s^{(i)}$ denotes the corresponding style label. The objective of the task is to generate (via a generator $g$) the output with the target style conditioned on $s$
while preserving most of the semantics of the source $x$. In other words, $\hat{x} = g(x, s)$ should have style $s$ and the semantics of $x$. We define the style as a binary attribute such that $s \in \{0, 1\}$, however, it can be easily extended to a multi-class setting.

\subsection{Generator}
For our generator, we fine-tune a large-scale language model GPT-2 \cite{radford2019language}. GPT-2 is pre-trained on large corpora and can be fine-tuned to generate fluent and coherent outputs for a variety of language generation tasks \citep{wolf2019transfertransfo}.
Since GPT-2 is a unidirectional language model, we reformulate the conditional generation task as a sequence completion task. Namely, as input to the generator, we concatenate the original sentence with a special token which indicates the target style.
The sequence following the style token is our output.

\begin{figure}[bt!]
    \centering
    \includegraphics[width=8cm]{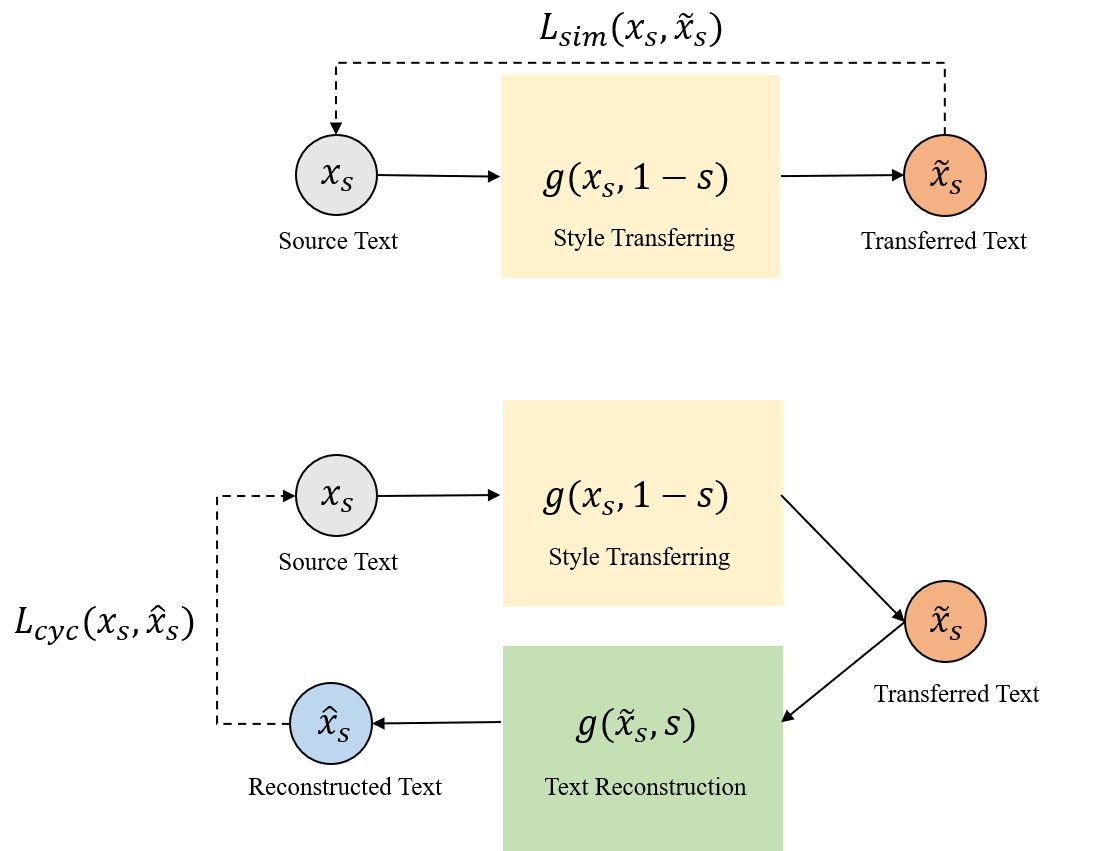}
    \caption{SIM Loss v.s. Cycle-Consistency Loss}
    \label{fig:sim}
\end{figure}

\subsection{Reward Functions}
We use four reward functions to control the quality of the system outputs. The  quality of the outputs is assessed in three ways: style transfer accuracy, content preservation, and fluency. We attend to each of these factors with their respective rewards. Here we denote the input text $x$ having style $s$ by $x_s$, and denote the output by $\tilde{x}_s$, i.e., $\tilde{x}_s = g(x_s, 1 - s)$.

\paragraph{Rewards for Style Transfer Accuracy}
We use a style classifier to provide the supervision signal to the generator with respect to the style transfer accuracy.
The min-max game between the generator $g$ and the classifier $f_{cls}$ is:
\begin{equation}
    \begin{split}
        & \min_{\theta_g}\max_{\theta_{f_{cls}}}\mathbb{E}_{x_s}[\log(1-f_{cls}(g(x_s, 1 - s), 1 - s))] \\
    &+ \mathbb{E}_{x_s}[\log f_{cls}(x_s, s) + \log (1-f_{cls}(x_s, 1 - s))]. \\
    \end{split}
\end{equation}
The style transfer accuracy reward for the generator is the log-likelihood of the output being labeled as the target style:
\begin{equation}
\label{eq:cls}
    r_{cls}(\tilde{x}_{s}) = \log(f_{cls}(\tilde{x}_{s}, 1 - s)).
\end{equation}
Following prior work, we use the CNN-based classifier \citep{kim-2014-convolutional} 
$f_{cls}$, which takes both the sentence and the style label as input and its objective is to predict the likelihood of the sentence being coherent to the given style.

\paragraph{Rewards for Content Preservation}

To ensure that the system outputs still preserve the basic semantics of the source sentences, we use the pre-trained SIM model introduced in \citet{wieting2019simple,wieting-etal-2019-beyond} to measure the semantic similarity between the source sentences and system outputs. The SIM score for a sentence pair is the cosine similarity of its sentence representations. These representations are constructed by averaging sub-word embeddings. Compared to the cycle-consistency loss~\citep{ijcai2019-0711, dai2019style, pang2019unsup}, our method is more direct since it doesn't require a second-pass generation. It also has advantages over $n$-gram based metrics like BLEU~\citep{papineni2002bleu} since it is more robust to lexical changes and can provide smoother rewards.

In~\citet{wieting-etal-2019-beyond}, SIM is augmented with a length penalty to help control the length of the generated text. We use their entire model, \simlength, as the content preservation reward,
\begin{equation}
\label{eq:sim}
     r_{sim}(\tilde{x}_s) = {\textrm{LP}(x_s, \tilde{x}_s)}^\alpha \textrm{SIM}(x_s, \tilde{x}_s),
\end{equation}
where 
\begin{equation}
    \textrm{LP}(r, h) = e^{1 - \frac{min(|r|, |h|)}{max(|r|, |h|)} },
\end{equation}
and $\alpha$ is an exponential term to control the weight of the length penalty, which is set to 0.25.

We also use the cycle-consistency loss $L_{cyc}$ to bootstrap the training:
\begin{equation}
     L_{cyc}(\theta_g) = \mathbb{E}_{x_s}[{-\log(p_{g}(x_s|g(x_s, 1-s), s))}].
\end{equation}
Here, $p_g$ is the likelihood assigned by the generator $g$. This introduces two generation passes, i.e., $\tilde{x}_s = g(x, 1-s)$ and $\bar{x}_s = g(\tilde{x}_s, s)$ while SIM reward only requires one generation pass, as illustrated in Fig. \ref{fig:sim}.

\paragraph{Rewards for Fluency}

Style transfer accuracy rewards and content preservation rewards do not have a significant effect on the fluency of the outputs. Therefore, we again use the pre-trained GPT-2 model, but as a reward this time. To encourage the outputs to be as fluent as the source sentences, we define the fluency reward as the difference of the perplexity between the system outputs and source sentences:
\begin{equation}
    \label{eq:lang}
    r_{lang}(\tilde{x}_s) =  \textit{ppl}(x_s) - \textit{ppl}(\tilde{x}_s).
\end{equation}

Here, $\textit{ppl}$ denotes the length-normalized perplexity assigned by the language model fine-tuned on the training set.

As will be further discussed in Section~\ref{sub:adv}, we found that using the rewards mentioned above can still result in unnatural outputs. Therefore, we additionally use a LSTM-based \citep{hochreiter1997long} discriminator
$f_{adv}$ to provide a naturalness reward, whose job is to discriminate the system outputs and the real sentences, i.e., an adversarial discriminator. 
It constructs a min-max game with the generator:
\begin{equation}
\begin{split}
     & \min_{\theta_g}\max_{\theta_{f_{adv}}}\mathbb{E}_{x_s}[\log(1-f_{adv}(g(x_s, 1 - s)))] \\
    &+ \mathbb{E}_{x_s}[\log (f_{adv}(x_s))]. \\
\end{split}
\end{equation}
The naturalness reward is the log-likelihood of the outputs being classified as real sentences:
\begin{equation}
\label{eq:adv}
    r_{adv}(\tilde{x}_{s}) = \log(f_{adv}(\tilde{x}_{s})).
\end{equation}

\subsection{Learning}
\label{sub:learning}

The final corresponding loss term is: 
\begin{equation}
    L_{*}(\theta_g) = - \frac{1}{N}\sum_{i=1}^N r_{*}(\tilde{x}_s^{(i)}).
\end{equation}
Here, $N$ is the number of samples in the dataset.
To train the model, we use the weighted average of the losses defined in the previous section:
\begin{equation}
\label{eq:loss}
    \begin{split}
        L(\theta_g) &= \lambda_{cls} L_{cls}(\theta_g) + \lambda_{adv} L_{adv}(\theta_g) \\
     & + \lambda_{sim} L_{sim}(\theta_g) + \lambda_{lang} L_{lang}(\theta_g) \\
     & + \lambda_{rec} L_{rec}(\theta_g).
    \end{split}
\end{equation}
where $\lambda$ denotes the weight of the corresponding term.
The setting of $\lambda$ is chosen to make the training stable and have balanced style transfer accuracy and content preservation performance on the development set. 
$L_{rec}$ is the reconstruction loss, i.e., 
\begin{equation}
    L_{rec}(\theta_g) = \mathbb{E}_{x_s}[{-\log(p_{g}(x_s|x_s, s))}].
\end{equation}

We follow a two-stage training procedure. We first use the cycle-consistency loss $L_{cyc}$
to bootstrap the training and then fine-tune the model with the rewards we introduced above to improve the output quality. 

In the bootstrap stage, the objective function is
\begin{equation}
\label{eq:bootstrap}
    \begin{split}
        L_{boot}(\theta_g) &= \lambda_{cyc} L_{cyc}(\theta_g) + \lambda_{cls} L_{cls}(\theta_g) \\
     & + \lambda_{rec} L_{rec}(\theta_g)
    \end{split}
\end{equation}

We select the checkpoint with the highest mean of the style transfer accuracy and BLEU on the development set as the starting point for the second training stage.

In the second stage, the generator is optimized with Eq.~\ref{eq:loss}. The classifier $f_{cls}$ for $L_{cls}$ is pre-trained and the language model for $L_{lang}$ is fine-tuned on the training set. During training, the discriminator $f_{adv}$ for $L_{adv}$ is trained against the generator. $f_{cls}$ is fixed when trained on some datasets, while it is trained against the generator on others.
We select the checkpoint that has the style transfer accuracy and BLEU score similar to that from the first stage and the lowest perplexity on the development set.

Lastly, since gradients can not be propagated through the discrete samples, we use two approaches to circumvent this problem. For the content preservation reward (Eq.~\ref{eq:sim}) and fluency reward (Eq.~\ref{eq:lang}), we use the REINFORCE~\citep{williams1992simple} algorithm to optimize the model, 
\begin{equation}
    \begin{split}
    &\nabla_{\theta_g} \mathbb{E}_{\tilde{x}_s \sim p_g(\tilde{x}_s)}[r(\tilde{x}_s)]\\ &= \mathbb{E}_{\tilde{x}_s \sim p_g(\tilde{x}_s)} [\nabla_{\theta_g}\log{p_g(\tilde{x}_s)}r(\tilde{x}_s)]
\end{split}
\end{equation}
We approximate the expectation by greedy decoding and the log-likelihood is normalized by sequence length, i.e.,~$\frac{1}{L}\sum_{i=1}^L \log p_g(\tilde{w}_i)$, where $\tilde{w}_i$ denotes the $i$-th token of $\tilde{x}_s$ and $L$ is sequence length. For the style transfer accuracy reward (Eq.~\ref{eq:cls}) and naturalness reward (Eq.~\ref{eq:adv}), we use a different approach to generate a continuous approximation of the discrete tokens, which allows gradients to be back-propagated to the generator. Namely, taking the style classifier $f_{cls}$ as an example, we use the distribution $p_i$ of each token produced by the generator as the input of the classifier. This distribution is then multiplied by the classifier's word embedding matrix $W^{embed}$ to obtain a weighted average of word embeddings:
\begin{equation}
    \hat{w}_i = p_iW^{embed}
\end{equation}
Then, the classifier takes the sequence of $\hat{w}_i$ as its input.
We chose this method because it provides a token-level supervision signal to the generator,
while the REINFORCE algorithm provides sentence-level signals.

\section{Experiments}
\subsection{Datasets}
We evaluate our approach on three datasets for sentiment transfer with positive and negative reviews: Yelp review dataset, Amazon review dataset provided by \citet{li-etal-2018-delete},\footnote{\url{https://github.com/lijuncen/Sentiment-and-Style-Transfer}}
and the IMDb movie review dataset provided by \citet{dai2019style}.\footnote{\url{https://github.com/fastnlp/nlp-dataset}}

We also evaluate our methods on a formality style transfer dataset, Grammarly's Yahoo Answers Formality Corpus (GYAFC),\footnote{\url{https://github.com/raosudha89/GYAFC-corpus}}
introduced in \citet{rao-tetreault-2018-dear}. Although it is a parallel corpus, we treat it as an unaligned corpus in our experiments. In order to compare to previous work, we chose the \textit{Family \& Relationships} category for our experiments. 
Datasets statistics are shown in Table~\ref{tab:dataset}. 

\begin{table}
\centering
\small
\begin{tabular}{lllll}
\toprule 
\textbf{Dataset} & \textbf{Style} & \textbf{Train} & \textbf{Dev} & \textbf{Test}\\
\midrule 
\multirow{2}{*}{Yelp} & Positive & 266K & 2000 & 500 \\
& Negative & 177K & 2000 & 500 \\
\midrule 
\multirow{2}{*}{Amazon} & Positive & 277K & 985 & 500 \\
& Negative & 279K & 1015 & 500\\
\midrule
\multirow{2}{*}{IMDb} & Positive & 178K & 2000 & 1000 \\
& Negative & 187K & 2000 & 1000\\
\midrule
\multirow{2}{*}{GYAFC} & Formal & 52K & 2247 & 500 \\
& Informal & 52K & 2788 & 500\\
\bottomrule
\end{tabular}
\caption{\label{tab:dataset} Number of samples in the Train, Dev, and Test splits for each dataset in our experiments.
}
\end{table}

\begin{table}
\centering
\small
\addtolength{\tabcolsep}{-2pt}
\begin{tabular}{llllllll}
\toprule 
\textbf{Dataset} & \textbf{Eq.} & $\lambda_{cls}$ & $\lambda_{adv}$ & $\lambda_{sim}$ & $\lambda_{lang}$ &  $\lambda_{rec}$ &  $\lambda_{cyc}$ \\
\midrule 
\multirow{2}{*}{Yelp} & (10) & 2 & 0.5 & 20 & 2 & 0.1 & - \\
& (12) & 1 & - & - & - & 1 & 1.5 \\
\midrule 
\multirow{2}{*}{Amazon} & (10) & 2 & 0.5 & 20 & 2 & 1 & - \\
& (12) & 5 & - & - & - & 1 & 0.5 \\
\midrule
\multirow{2}{*}{IMDb} & (10) & 1 & 0.5 & 20 & 2 & 1 & - \\
& (12) & 1 & - & - & - & 1 & 1 \\
\midrule
\multirow{2}{*}{GYAFC} & (10) & 2 & 0.5 & 20 & 2 & 1 & - \\
& (12) & 1 & - & - & - & 1 & 1 \\
\bottomrule
\end{tabular}
\addtolength{\tabcolsep}{+2pt}
\caption{\label{tab:hyper} Hyperparameter setting of Eq.~\ref{eq:loss} and Eq.~\ref{eq:bootstrap} on each dataset. 
}
\end{table}

\subsection{Experimental Details}

Following previous work, we measure the style transfer accuracy using a  FastText\footnote{\url{https://fasttext.cc/}}~\citep{joulin2017bag} style classifier trained on the respective training set of each dataset. 
To measure content preservation, we use SIM and BLEU as metrics where self-SIM and self-BLEU are computed between the source sentences and system outputs, while ref-SIM and ref-BLEU are computed between the system outputs and human references when available. 
To measure the fluency we use a pre-trained GPT-2 model to compute the perplexity.\footnote{Note that we didn't fine-tune it on the training set}
Our generator, GPT-2, has 1.5 billion parameters, and we train on a GTX 1080 Ti GPU for about 12 hours.

The weights of the loss terms in Eq.~\ref{eq:loss} and Eq.~\ref{eq:bootstrap} are detailed in Table~\ref{tab:hyper}.
While during our experiments we found that there are other possible configurations which give higher scores with respect to the automatic evaluation metrics, as will be discussed in Section~\ref{sub:adv}, we also found that better performance in automatic evaluation doesn't always entail better performance in human evaluation.
Therefore, we also manually checked the quality of the transferred texts on development set when we chose the value of the hyperparameters. 

We compare our model with several state-of-the-art methods: DeleteAndRetrieve (D\&R)~\cite{li-etal-2018-delete}, B-GST~\cite{sudhakar2019transforming}, Cycle-Multi~\cite{dai2019style}, Deep-Latent~\cite{He2020A}, Tag\&Gen~\cite{madaan-etal-2020-politeness}, and DualRL~\cite{ijcai2019-0711}. 
We also compare our final model, \textbf{\model}(\textbf{Dir}ect-\textbf{R}eward), with the model only trained with the first stage (\modelcycle) as mentioned in Section \ref{sub:learning}.

\subsection{Adversarial Examples}
\label{sub:adv}

\begin{table}[t!]
\centering
\small
\addtolength{\tabcolsep}{-2pt}
\begin{tabular}{llccc}
\toprule
\textbf{Dataset}&\textbf{Model} & \textbf{Acc} & \textbf{PPL} & \textbf{BLEU} \\
\midrule
\multirow{2}{*}{Yelp} & \modelcycle & 91.7 & 392 & 18.7 \\
& \textsc{DirR}-\textsc{Yelp}-\textsc{Adv} & 95.2 & 353 & 20.7  \\
\midrule
\multirow{2}{*}{Amazon} & \model & 62.2 & 205 & 30.1  \\
& \textsc{DirR}-\textsc{Amazon}-\textsc{Adv} & 83.2 & 228 & 29.0 \\
\bottomrule
\end{tabular}
\addtolength{\tabcolsep}{+2pt}
\caption{ \label{tab:adv}
Adversarial Results.
\textbf{\textsc{DirR}-\textsc{Yelp}-\textsc{Adv}} and \textbf{\textsc{DirR}-\textsc{Amazon}-\textsc{Adv}} denote the models which generate adversarial examples. \textbf{Acc} denotes the style transfer accuracy, \textbf{PPL} denotes the perplexity, \textbf{BLEU} is computed between the human references and system outputs. 
}
\end{table}

Yelp and Amazon are arguably the most frequently used datasets for the sentiment transfer task. In our experiments, we found that the automatic evaluation metrics can be tricked on these datasets. 
Table~\ref{tab:adv} shows the performance of the models which generate adversarial examples. Upon identifying these risks, we propose several design options that can effectively mitigate these problems.

\paragraph{Yelp Dataset}For the Yelp dataset, when trained without the adversarial discriminator $f_{adv}$ and the fluency reward,
our model (\textsc{DirR}-\textsc{Yelp}-\textsc{Adv}) is able to discover a trivial solution which receives high automatic evaluation scores: injecting a word that carries strong sentiment at the beginning of the output, and making minimum changes (if any) to the source sentences, as illustrated in Table~\ref{tab:comparison}. This obviously does not meet the objective of content-preserving sentiment transfer and is easily detectable for humans. In fact, after we manually removed the first word from each of the output sentences, the transfer accuracy dropped from 95.2 to 58.4. To address this problem, we introduced an auxiliary discriminator $f_{adv}$ as we discussed above to penalize the trivial outputs since they can be easily captured by the discriminator. On the other hand, the output perplexity is not sensitive enough to this local feature so using the fluency reward alone is not sufficient. Our final model has much more stable performance when the first word of its output sentences is removed, experiencing only a small drop of the style transfer accuracy from 94.2 to 88.2. 

\paragraph{Amazon Dataset} For the Amazon dataset, we found that the style classifier $f_{cls}$ needs to be updated during the training to prevent the model exploiting the data imbalance problem of the dataset.
Namely, in the Amazon dataset some categories of products appear mostly in negative or positive reviews. In Table~\ref{tab:freq}, we show the word frequency of \textit{game} and \textit{phone} in both negative and positive reviews. In the original dataset, \textit{game} mostly appears in negative reviews while \textit{phone} mostly appears in positive reviews. Therefore, without any prior knowledge, it is very likely that these words will be used as informative features by the sentiment classifier, which makes its predictions unreliable.\footnote{Notice that the style classifier only achieves 43 accuracy on the human references.}

\begin{table}[t!]
\centering
\small
\begin{tabular}{lcccc}
\toprule
\multirow{2}{*}{\textbf{Model}} & \multicolumn{2}{c}{"game"} &  \multicolumn{2}{c}{"phone"}\\\cmidrule{2-3} \cmidrule{4-5}
& Pos. & Neg. & Pos. & Neg. \\
\midrule
Train & 58 & 7548 & 8947 & 2742 \\
Test & 0 & 10 & 20 & 6 \\
Human & 1 & 10 & 18 & 6 \\
B-GST & 55 & 0 & 13 & 44 \\
Tag\&Gen & 69 & 0 & 14 & 5 \\
\model & 26 & 0 & 19 & 45 \\
\textsc{DirR}-\textsc{Amazon}-\textsc{Adv} & 291 & 0 & 190 & 4 \\
\bottomrule
\end{tabular}
\caption{\label{tab:freq}
Frequencies of words in the Amazon Dataset that appear often enough in specific classes to erroneously cause the classifier to make incorrect predictions. \textbf{Pos.} denotes the positive sentences, \textbf{Neg.} denotes the negative sentences. 
}
\end{table}

\begin{table}[h!]
\centering
\small
\addtolength{\tabcolsep}{-2pt}
\begin{tabular}{ll}
\toprule
\textbf{Model} & \textbf{Text}\\
\midrule
Source & don t waste your time or money on these jeans . \\
Adv & don t need your time or money on these \textbf{phones} . \\
\midrule
\multirow{2}{*}{Source} & i made beef bolognese in the oven and it turned  \\
& out wonderfully . \\
\multirow{2}{*}{Adv} & i made beef bolognese in the \textbf{game} and it turned \\
&  out wonderfully . \\
\midrule
Source & this one does the job i need it for ! \\
Adv & this \textbf{game} does the job i need it for !\\
\bottomrule
\end{tabular}
\addtolength{\tabcolsep}{+2pt}
\caption{\label{tab:examples}
Adversarial examples received high style transfer accuracy scores on Amazon Dataset. Adv denotes the adversarial examples generated by \textsc{DirR}-\textsc{Amazon}-\textsc{Adv}.
}
\end{table}

When our second-stage model is trained with the fixed style classifier, it (\textsc{DirR}-\textsc{Amazon}-\textsc{Adv}) learns to exploit this dataset bias by changing the nouns in the original sentences to \textit{game} or \textit{phone}, which achieves better transfer accuracy. We list some examples in Table~\ref{tab:examples}. \textsc{DirR}-\textsc{Amazon}-\textsc{Adv} generated 291 \textit{game} in 500 positive reviews, which obviously changes the semantics of the source sentences. In order to show that this phenomenon is independent to the classifier architecture, we additionally fine-tuned a BERT-based~\citep{devlin-etal-2019-bert} classifier, which yielded 51.3, 57.6, 70.4 accuracy on human references, \model, \textsc{DirR}-\textsc{Amazon}-\textsc{Adv} respectively, showing the same pattern of the fastText classifier. 
We notice that some two-stage models~\citep{li-etal-2018-delete,sudhakar2019transforming,madaan-etal-2020-politeness} and other methods~\citep{yang2018unsupervised, ijcai2019-0711} also use a fixed classifier or use words with unbalanced frequencies in different styles as important features, which means that their methods may face the same risk. 
While \citet{li-etal-2018-delete} has pointed out this data imbalance problem of the Amazon dataset, we further demonstrate that a strong generator can even use this discrepancy to trick the automatic metrics. 
We are able to mitigate this problem by updating the style classifier during the training, and in Table~\ref{tab:freq}, \model is more robust to the data imbalance problem compared to other methods.

\begin{table}[bt!]
\small
\centering
\begin{tabular}{lcccc}
\toprule
\textbf{Model} & \textbf{Acc} & \textbf{PPL} & \textbf{r-BLEU} & \textbf{s-BLEU}\\
\midrule
\multicolumn{5}{c}{Yelp} \\
\midrule
 D\&R & 89.0 & 362 & 10.1 & 29.1  \\
 B-GST & 86.0 & \textbf{269} & 14.5  & 35.1 \\
 Cycle-Multi & 87.6 & 439 & 19.8 & \textbf{55.2}  \\
Deep-Latent & 86.0 & 346 & 15.2 & 40.7  \\
Tag\&Gen & 88.7 & 355 & 12.4 & 35.5 \\
\modelcycle & 91.7 & 392 & 18.7  & 51.2  \\
\model & \textbf{94.2} & 292 & \textbf{20.7}  & 52.6 \\ 
Copy & 4.1 & 204 & 22.5  & 100.0  \\
Human & 70.7 & 236 & 99.3 & 22.5 \\
\midrule
\multicolumn{5}{c}{Amazon} \\
\midrule
 D\&R & 50.0 & 233 & 24.1 & 54.1  \\
 B-GST & 60.3 & \textbf{197} & 20.3  & 44.6  \\
 Tag\&Gen & \textbf{79.9} & 312 & 27.6  & \textbf{62.3} \\
 \modelcycle & 68.4 & 374 & 29.0 & 60.6 \\
 \model & 62.2 & 205 & \textbf{30.1} & 61.3 \\ 
 Copy & 21.1 & 218 & 40.0 & 100.0\\
 Human & 43.0 & 209 & 100.0 & 40.0  \\
\midrule
\multicolumn{5}{c}{IMDb} \\
\midrule
 Cycle-Multi & 77.1 & 290 & N/A  & \textbf{70.4} \\
 \modelcycle & 80.5 & 253 & N/A  & 64.3 \\
 \model & \textbf{83.2} & \textbf{210} & N/A & 64.2  \\ 
 Copy & 5.3 & 147 & N/A  & 100.0  \\
\midrule
\multicolumn{5}{c}{GYAFC} \\
\midrule

 D\&R & 51.2 & 226 & 14.4& 27.1 \\
 DualRL & 62.0 & 404 & 33.0  & 50.8 \\
 \modelcycle & \textbf{76.2}& 162 & 44.1 & \textbf{66.5}\\
 \model & 71.8 & \textbf{145} & \textbf{46.3} & 59.9 \\ 
 Copy & 15.8 & 147 & 41.5 & 98.5\\
 Human & 84.5 & 137 & 97.8 & 21.5 \\
\bottomrule
\end{tabular}
\caption{\label{tab:result}
Automatic Evaluation. Acc is the accuracy of the sentiment classifier. PPL is the perplexity assigned by the GPT-2 language model. r-BLEU is the BLEU score between the human references and system outputs. s-BLEU is the BLEU score between the source sentences and system outputs. Copy is an oracle which copies the source sentences as outputs. Human denotes the human references. 
}
\end{table}

\subsection{Automatic Evaluation}

The automatic evaluation results are shown in Table~\ref{tab:result}. We report the performance of the previous methods based on the outputs they provided for fair comparison and omit those whose results are not available. 

We have the following observations of the results.
First, compared to our base model (\modelcycle), the model trained with our proposed rewards has higher fluency, while remains the same level of content preservation. It indicates that SIM score is as effective as cycle-consistency loss for content preservation and our fluency reward can effectively improve the output fluency. Secondly, there exists a trade-off among the style transfer accuracy, content preservation and language fluency. While our model does not outperform the previous methods on all of the metrics, it is able to find a better balance of the different metrics.

\subsection{Human Evaluation}

\begin{table}[bt!]
\centering
\small
\begin{tabular}{llcccc}
\toprule
\textbf{Dataset} & \textbf{Model} & \textbf{Style} & \textbf{Flu.} & \textbf{Con.} & \textbf{Mean}\\
\midrule
\multirow{3}{*}{Yelp} & Cycle & 2.24 & 0.62 & 1.97 & 2.02\\
& B-GST &\textbf{2.42}& 0.64 & 2.02 & 2.12\\
& \model & \textbf{2.42} & \textbf{0.66} & \textbf{2.04} & \textbf{2.14} \\ 
\midrule
\multirow{3}{*}{Amazon} & Tag\&Gen & 1.98 & 0.87 & 1.95 & 2.19\\
& B-GST &2.04& \textbf{0.89} & 1.77 & 2.16\\
& \model* & \textbf{2.09} & 0.87 & \textbf{2.10} &\textbf{2.26}\\ 
\midrule
\multirow{3}{*}{GYAMC} & D\&R & N/A & 0.40 & 2.13 & 1.66\\
& DualRL & N/A & 0.51 & 2.23 & 1.88\\
& \model* & N/A & \textbf{0.70} & \textbf{2.34} & \textbf{2.22} \\ 
\bottomrule
\end{tabular}
\caption{\label{tab:human}
Human Evaluation. \textbf{Style} denotes style transfer accuracy, \textbf{Flu.} denotes fluency, \textbf{Con.} denotes content preservation. \textbf{Mean} denotes the average of the metrics where the fluency scores are scaled up to be consistent with other scores. *: significantly better than other systems ($p < 0.01$) according to the mean score.
}
\end{table}

\begin{table*}[t]
\centering
\addtolength{\tabcolsep}{-2pt}
\begin{tabular}{llcc}
\toprule
\textbf{Model} & \textbf{Text} & \textbf{self-BLEU} & \textbf{self-SIM} \\
\midrule
Source & this was my first stop in looking for a wedding dress .
& 100.0 & 100.0 \\ 
\modelbleu & \textbf{great} this was my first stop in looking for a wedding dress .
& 91.2 & 95.2 \\ 
\model & this was my best stop in looking for a wedding dress .
& 64.8 & 81.9 \\ 
\midrule
source &  just a frozen patty cooked like a home one .
& 100.0 & 100.0 \\ 
\modelbleu & \textbf{great} a frozen patty cooked like a home one .
& 88.0 & 94.6 \\ 
\model & just a great patty cooked like a home one .
& 70.7 & 88.5 \\ 
\midrule
\multirow{2}{*}{source} &   wendy 's has been know to be cheap with their drink 
& \multirow{2}{*}{100.0} & \multirow{2}{*}{100.0}\\ 
& refills for years . & & \\
\multirow{2}{*}{\modelbleu} & \textbf{great} wendy 's has been know to be cheap with their drink 
& \multirow{2}{*}{93.0} & \multirow{2}{*}{97.5}\\ 
& refills for years . & & \\
\model & wendy 's has been great with their drink refills for years .
& 57.2 & 84.9 \\ 
\bottomrule
\end{tabular}
\addtolength{\tabcolsep}{+2pt}
\caption{\label{tab:comparison}
Comparison of using SIM and BLEU as the content preservation reward. Samples are from the Yelp dataset. 
The metrics self-BLEU and self-SIM are calculated between the source sentences and system outputs.}
\end{table*}

We conducted human evaluation on Yelp, Amazon and GYAFC datasets evaluating the style transfer accuracy, content preservation, and fluency separately. 
The first two aspects are rated with range 1 - 3 while the fluency is rated with range 0 - 1.
We randomly select 100 candidates and compare the outputs of different systems. 
 We use Amazon Turk\footnote{\url{https://www.mturk.com/}} for human evaluation. Each candidate is rated by three annotators and we report the average scores here. We did not evaluate the style transfer accuracy  for the GYAMC dataset since it is difficult for human annotators to accurately capture the difference between formal and informal sentences. The results of our human evaluations are shown in Table~\ref{tab:human}. We additionally report the sample-wise mean score of the metrics where the fluency scores are scaled up to be consistent with other scores. Our model achieves better overall performance when considering all three evaluation metrics on each dataset. 

Interestingly, we found that the automatic metrics for both the style transfer accuracy and content preservation do not accurately reflect performance as measured by human evaluation. 
For example, on the Amazon dataset, although Tag\&Gen~\citep{madaan-etal-2020-politeness} achieves significantly higher style transfer accuracy based on the automatic metric, our model achieves better performance based on the human evaluation. 
This phenomenon suggests that the importance of our findings discussed in Section~\ref{sub:adv}, that strong neural models can potentially exploit the weaknesses of the automatic metrics.

\section{Analysis}
\label{sec:analysis}
\begin{table}
\centering
\small
\addtolength{\tabcolsep}{-2pt}
\begin{tabular}{lcccc}
\toprule
\textbf{Model} & \textbf{Acc} & \textbf{PPL} & \textbf{s-BLEU} & \textbf{s-SIM}\\
\midrule
\modelcycle & 91.7 & 392 & 51.2 & 76.2\\
\modelsim & 92.1 & 348 & 51.4 & 79.8\\
\modelbleu & 91.3 & 315 & \textbf{59.4} & \textbf{81.8} \\
\model & \textbf{94.2} & \textbf{292} & 52.6 & 81.6 \\ 
\bottomrule
\end{tabular}
\addtolength{\tabcolsep}{2pt}
\caption{\label{tab:ablation}
Ablation and Comparative Study on Yelp Dataset. 
Acc is the accuracy of the sentiment classifier. PPL is the perplexity assigned by the GPT-2 language model.
self-BLEU (s-BLEU) and self-SIM (s-SIM) are computed between the source sentences and outputs.
}
\end{table}

We next show an ablation study, demonstrating the effectiveness of the content preservation  and fluency rewards in \model, and how SIM can be used to replace the cycle-consistency loss. We also compare using BLEU versus using SIM as a content-preservation reward, finding that using BLEU results in reduced performance, unstable training, and artifacts in the outputs, which makes the results less natural than the results of the model trained with SIM score.

To illustrate that training with SIM can replace the cycle-consistency loss for content preservation, we fine-tuned \modelcycle on SIM to produce a new model, \modelsim. The difference between \model and \modelsim is that the former is additionally trained with our fluency rewards. The results are shown in Table \ref{tab:ablation}, and show two main trends. First, we see that \modelsim has better fluency and content preservation performance than \modelcycle, which shows that the cycle-consistency loss can be replaced by SIM score for content preservation. 
Second, \model has better fluency than \modelsim, showing the effectiveness of our fluency rewards.

We next investigate the effectiveness of using SIM as a reward instead of BLEU. To do this, we train a model, \modelbleu, which uses BLEU as the content reward and report the results in Table~\ref{tab:ablation}. The results show that using BLEU has larger content preservation as measured by BLEU, but has similar performance  when measured by SIM. However, performance on the style transfer accuracy and fluency decreases. We hypothesize that this is because using SIM as a reward gives the model more freedom, allowing the model to have more balanced performance since there is less pressure to copy $n$-grams. We also observe more adversarial examples in the outputs of \modelbleu. As discussed in Section \ref{sub:adv}, these adversarial examples are generated by injecting a word carrying strong sentiment at the beginning of the output. The model trained with BLEU is more likely to generate these outputs as it will try to avoid breaking up the $n$-grams in the source sentences, allowing for a higher BLEU reward. Examples of this behavior is shown in
Table \ref{tab:comparison}. Notice that the \modelbleu samples start with the word \textit{great}, which is enough to often fool the classifier, but are unnatural.

\section{Related Work}
A main line of work \citep{shen2017style, hu2017toward, fu2018style, xu2018unpaired, john-etal-2019-disentangled} for text style transfer aims to model the conditional distribution of the data with the encoder-decoder architecture. Due to the lack of parallel corpora, inductive biases are designed to make the generation conditioned on both source sentences and specific styles such that the model can rewrite the source texts with the target style while still preserve the content information of the source texts.

Efforts are also made to design training objectives to improve performance. For example, Back-translation \citep{DBLP:journals/corr/abs-1808-07894, prabhumoye-etal-2018-style}, denoising auto-encoding \citep{lample2018multipleattribute} and the cycle-consistency loss \citep{ijcai2019-0711, dai2019style, pang2019unsup} have been shown effective for improving the model performance. \citet{li-etal-2018-delete} proposes a retrieve-based pipeline, which contains three stages, namely, delete, retrieve and generate. \citet{sudhakar2019transforming} extends this pipeline by using GPT~\citep{radford2018improving} as the generator. 
Compared to these methods, we propose a more direct and effective approach to encourage semantic-preserving transfer by directly measuring the semantic similarity of the source texts and system outputs.

Recently, other works have been proposed for unsupervised text style transfer~\citep{jin2019imat, lai2019multiple, wu-etal-2019-hierarchical-reinforced, Li_Li_Zhang_Li_Zheng_Carin_Gao_2020}. \citet{He2020A} proposes a probabilistic view which models the non-parallel data from two domains as a partially observed parallel corpus.
\citet{madaan-etal-2020-politeness} proposes a tag-and-generate pipeline, which firstly identifies style attribute markers from the source texts, then replaces them with a special token, and generates the outputs based on the tagged sentences. \citet{zhou-etal-2020-exploring} focuses on exploring the word-level style relevance which is assigned by a pre-trained style classifier. They propose a reward for content preservation which is based on the weighted combination of the word embeddings of the source texts and system outputs. Compared to this reward, our proposed content reward is specifically designed for semantic similarity and pre-trained on large corpora, which makes it more robust across different datasets.  

\section{Conclusion}

In this paper, we propose a direct approach of improving content preservation for text style transfer by leveraging a semantic similarity metric as the content reward. Our proposed rewards that target different aspects of the output quality, enable our model to have strong performance on both automatic and human evaluation.
Recently, several semantic similarity metrics~\citep{zhao-etal-2019-moverscore,sellam-etal-2020-bleurt,gao-etal-2020-supert} based on pre-trained language models, have shown promising results. 
Introducing these metrics in our proposed method as the content preservation reward may lead to even further improvements.  

Moreover, we identify several problems in the commonly used automatic evaluation metrics and datasets. We propose several practical strategies to mitigate these problems, which makes these metrics more effective rewards for model training.
Considering the weaknesses of the automatic metrics presented in this work, we believe that more rigorous discussion and investigation of the criteria for successfully transferring style is essential.
Since existing work mostly relies on model-based metrics to determine the success of style transfer models, adversarial methods could be introduced to make the model-based metrics more robust and faithful indicators of style transfer. These improved evaluation models would then be beneficial for both learning and evaluating new approaches for style transfer.
 
\bibliography{anthology,custom}
\bibliographystyle{acl_natbib}

\end{document}